\documentclass[conference]{IEEEtran}
\IEEEoverridecommandlockouts
\usepackage{cite}
\usepackage{amsmath,amssymb,amsfonts}
\usepackage{algorithmic}
\usepackage{graphicx}
\usepackage{textcomp}
\usepackage{xcolor}
\usepackage{booktabs}
\usepackage{multirow}
\usepackage{comment}
\usepackage{tabularx}
\usepackage{subcaption}
\def\BibTeX{{\rm B\kern-.05em{\sc i\kern-.025em b}\kern-.08em
    T\kern-.1667em\lower.7ex\hbox{E}\kern-.125emX}}
\begin{document}

\title{DF-ACBlurGAN: Structure-Aware Conditional Generation of Internally Repeated Patterns for Biomaterial Microtopography Design\\
}

\makeatletter 
\newcommand{\linebreakand}{%
  \end{@IEEEauthorhalign}
  \hfill\mbox{}\par
  \mbox{}\hfill\begin{@IEEEauthorhalign}
}
\makeatother 
\author{\IEEEauthorblockN{Rongjun Dong}
\IEEEauthorblockA{\textit{University of Nottingham} \\
\textit{School of computer science}\\
Nottingham, UK \\
psxrd3@nottingham.ac.uk}
\and
\and
\IEEEauthorblockN{Xin Chen}
\IEEEauthorblockA{\textit{University of Nottingham} \\
\textit{School of computer science}\\
Nottingham, UK \\
Xin.Chen@nottingham.ac.uk}
\and
\IEEEauthorblockN{Morgan R Alexander}
\IEEEauthorblockA{\textit{University of Nottingham} \\
\textit{School of Pharmacy}\\
Nottingham, UK \\
Morgan.Alexander@nottingham.ac.uk}
\and
\and
\IEEEauthorblockN{Karthikeyan Sivakumar}
\IEEEauthorblockA{\textit{University of Nottingham} \\
\textit{School of Medicine}\\
Nottingham, UK \\
}
\and
\IEEEauthorblockN{Reza Omdivar}
\IEEEauthorblockA{\textit{University of Nottingham} \\
\textit{Digital Research Service}\\
Nottingham, UK \\
}
\and
\IEEEauthorblockN{David A Winkler}
\IEEEauthorblockA{\textit{Monash University} \\
\textit{La Trobe University}\\
Melbourne, AU \\
David.Winkler@monash.edu}
\and
\IEEEauthorblockN{Grazziela Figueredo}
\IEEEauthorblockA{\textit{University of Nottingham} \\
\textit{School of Medicine}\\
Nottingham, UK \\
G.Figueredo@nottingham.ac.uk}
}

\maketitle

\begin{abstract}
Learning to generate images with internally repeated and periodic structures poses a fundamental challenge for machine learning and computer vision models, which are typically optimised for local texture statistics and semantic realism rather than global structural consistency. This limitation is particularly pronounced in applications requiring strict control over repetition scale, spacing, and boundary coherence, such as microtopographical biomaterial surfaces. In this work, biomaterial design serves as a use case to study conditional generation of repeated patterns under weak supervision and class imbalance. We propose DF-ACBlurGAN, a structure-aware conditional generative adversarial network that explicitly reasons about long-range repetition during training. The approach integrates frequency-domain repetition scale estimation, scale-adaptive Gaussian blurring, and unit-cell reconstruction to balance sharp local features with stable global periodicity. Conditioning on experimentally derived biological response labels, the model synthesises designs aligned with target functional outcomes. Evaluation across multiple biomaterial datasets demonstrates improved repetition consistency and controllable structural variation compared to conventional generative approaches.
\end{abstract}

\begin{IEEEkeywords}
biomaterial surface topography, periodic pattern synthesis, DF-ACBlurGAN, Fourier analysis, GANs
\end{IEEEkeywords}

\section{Introduction}

Surface micro-topographies in biomaterials can modulate cellular behaviours such as macrophage polarisation and bacterial biofilm formation~\cite{b1,b2,b3}. High-throughput platforms have enabled systematic screening of cellular responses to 2,176 distinct combinatorial surface designs~\cite{b4,b5}. Each design is constructed by periodically repeating a geometric \emph{unit cell}, composed of a small set of basic shapes, across the chip surface (Fig.~\ref{fig:topochip_unitcell})~\cite{b8}. Analysis of these datasets has revealed quantitative associations between topographical descriptors and observed biological outcomes~\cite{b6,b7,b5}. However, because the designs were generated through pseudo-random combinations of a limited shape vocabulary, it remains unclear whether improved outcomes could be achieved through data-driven design optimisation. Advancing towards such intelligent design introduces several challenges. Existing libraries rely on a restricted set of primitive shapes, limiting structural diversity and novelty. Moreover, the biological response measurements are long-tailed and obtained by discretising continuous measured cellular responses, leading to class imbalance and ambiguous class boundaries that complicate learning-based modelling~\cite{b9,b10}.

\begin{figure}[t]
    \centering
    \includegraphics[width=\linewidth]{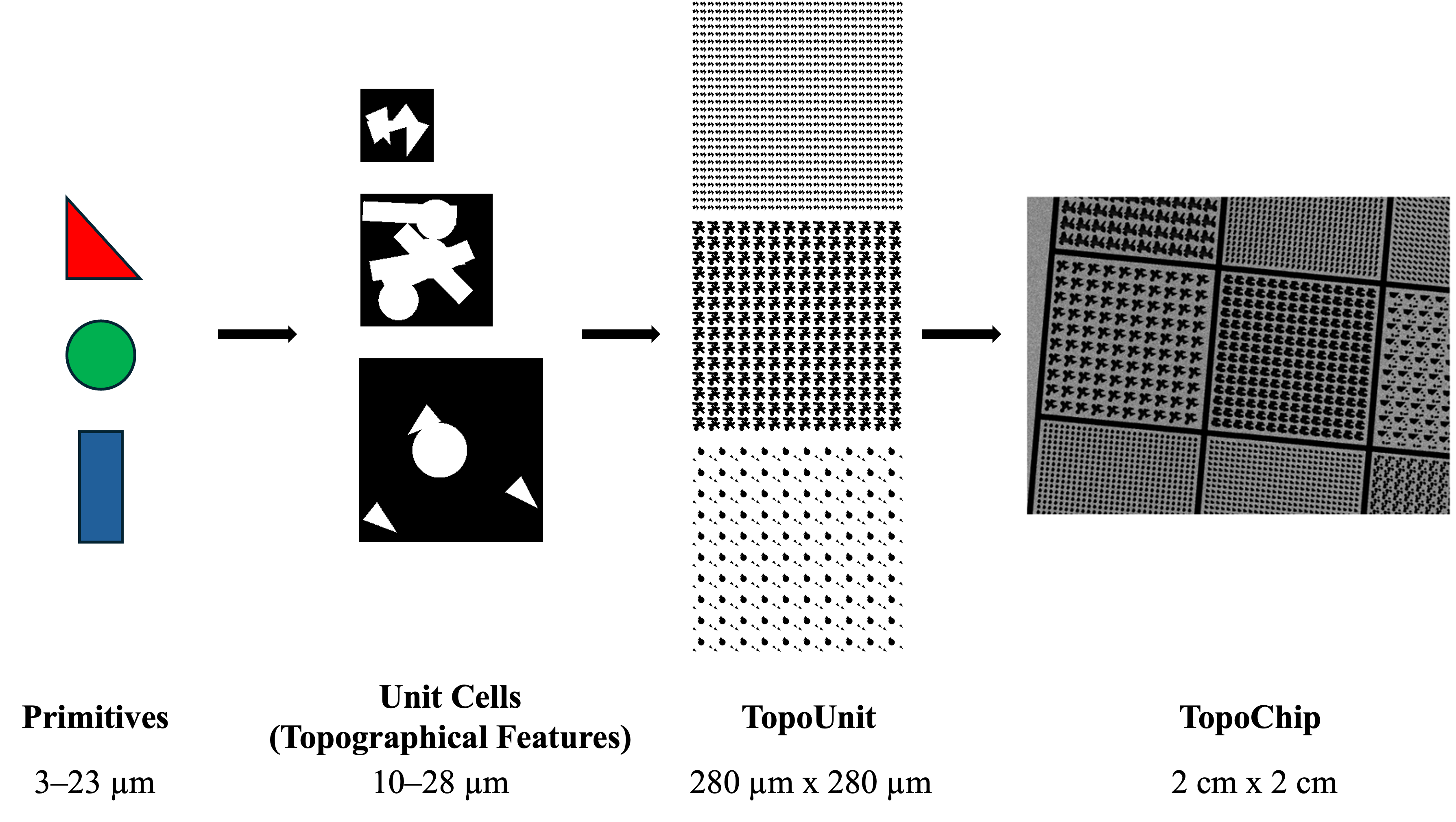}
    \caption{
    Topographical designs and their fabrication into a TopoChip used in high-throughput biological screening.
    }
    \label{fig:topochip_unitcell}
\end{figure}

Beyond data limitations, generating repeated topographies made of repeated patterns introduces additional challenges. Effective designs must preserve smooth and well-defined local feature boundaries, maintain consistent
global repetition, and respect biologically relevant spacing between repeated structures.
Naively generating a single unit cell and tiling it ignores boundary effects, spatial variability, and inter-unit spacing, all of which are known to influence biological behaviour\cite{b1,b2,b3}.
From a generative modelling perspective, these requirements introduce a tension between preserving sharp local features and suppressing high-frequency artefacts that can destabilise global repetition during adversarial training.
In practice, adversarial objectives tend to amplify such high-frequency artefacts, motivating the need for structure-aware mechanisms that stabilise feature boundaries while respecting repetition scale.
Such constraints necessitate explicit reasoning about repetition scale and structural organisation at the image level, rather than relying on infinite periodic repetition or purely local texture statistics. Microtopographical design is inherently a conditional generation problem.
Measured biological responses define functional targets,
such as promoting or inhibiting bacterial attachment or modulating macrophage behaviour.
Generative models must therefore be conditioned on experimentally observed response labels to produce designs that are structurally valid and target desirable biological outcomes.

Generative models offer a potential solution by enabling data-driven synthesis of novel designs~\cite{b11,b12,b13}.
However, most existing conditional generative approaches are primarily optimised for perceptual realism and semantic conditioning, as exemplified by large-scale GANs and diffusion models that focus on local texture statistics and class-level semantics~\cite{b14,b15,b16,b17}.
Explicit mechanisms for reasoning about global periodicity or internal structural repetition are rarely incorporated into the generative process. In particular, commonly used convolutional generators implicitly assume local stationarity, making it unclear whether they can reliably model long-range repetition and spacing without additional structural guidance. This raises questions about whether standard convolution-dominated architectures are sufficient for repeated-pattern generation at the scale and regularity required here.
Although frequency-domain characteristics of generated images have been analysed in prior work~\cite{b18,b19}, the problem of enforcing long-range repetition consistency in large, patterned surfaces remains largely under explored in current generative modelling literature.
Such limitations are especially critical for biomaterial surfaces and other areas that require the creation of repeated pattern designs that are context-aware, where internal repetition and global organisation directly influence a desired endpoint.

In this work, we propose Dynamic FFT Conditional Blurring Generative Adversarial Network (\textbf{DF-ACBlurGAN}), a structure-aware conditional generative architecture to synthesise images of designs with internally repeated and periodic structures under class imbalance and weak semantic supervision.
The model explicitly integrates frequency-domain pattern estimation, scale-adaptive Gaussian blurring, and unit-cell reconstruction into the adversarial training loop. We evaluate the framework across three biomaterial datasets involving bacterial biofilm formation and macrophage response. Generative quality is validated using complementary quantitative metrics and downstream surrogate models.

\section{Methods}

\begin{figure*}[t]
    \centering
    \includegraphics[width=\linewidth]{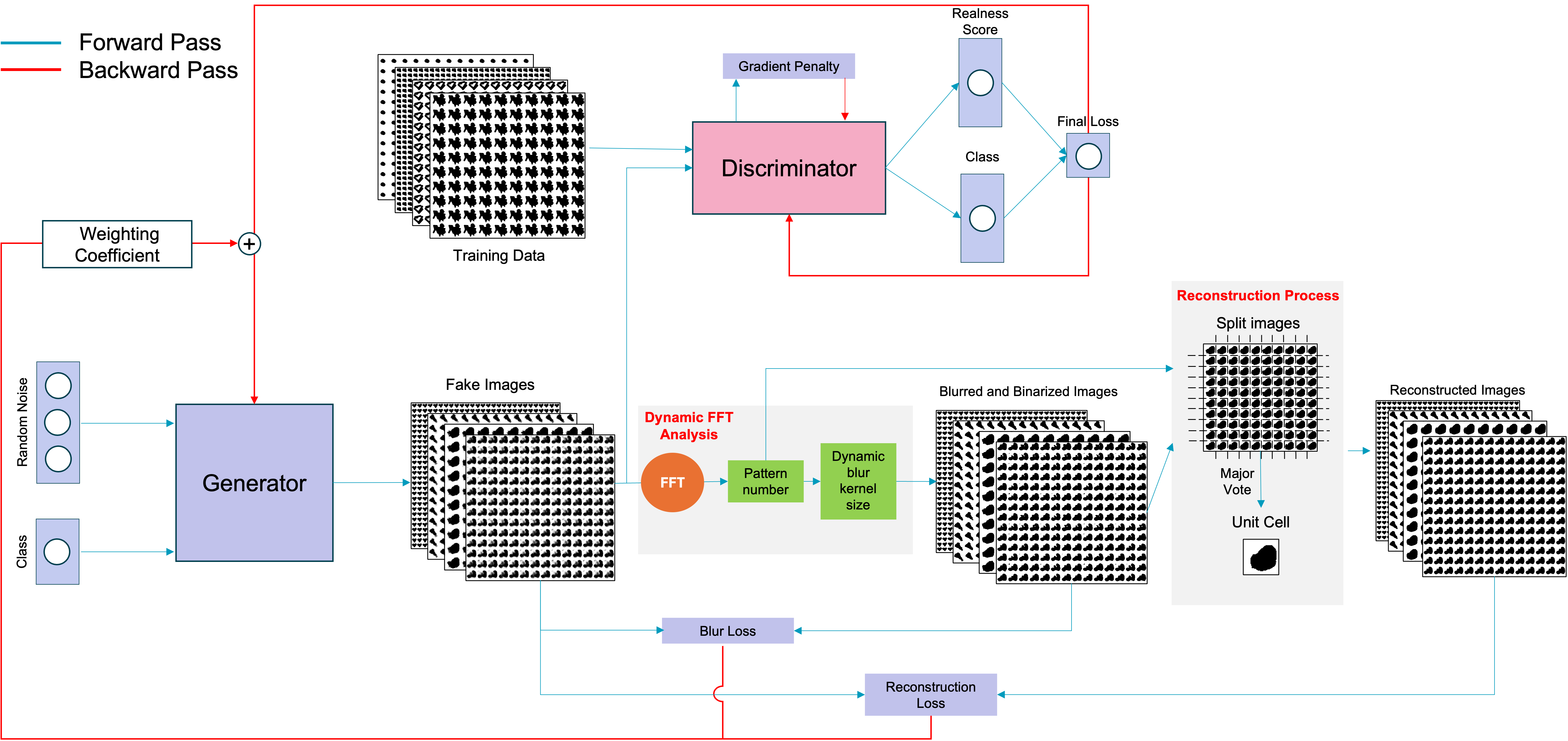}
    \caption{The DF-ACBlurGAN architecture.}
    \label{fig:framework}
\end{figure*}

DF-ACBlurGAN (Fig.~\ref{fig:framework}) was designed for images characterized by internal repetition of structural patterns. 
Image generation is guided by an existing 2,176 images dataset coupled with their labels (cellular responses) but also by inferred structural cues. 
Given a noise vector and a class label, the generator produces an initial topographical design. 
This intermediate output is then analysed to infer the number of repeating patterns within the image. The estimated structural scale is used to control a series of refinement operations, including scale-adaptive blurring and unit-cell-based reconstruction applied to the generated image. This enforces repetition consistency prior to loss evaluation. The adversarial training objective incorporates multiple loss functions that allow for the generator to learn representations consistent with semantic conditioning and internal structural organization, as further detailed next.

\paragraph{Conditional generative backbone}\label{AA}
DF-ACBlurGAN builds upon a class-conditional Wasserstein generative adversarial network with gradient penalty (WGAN-GP)~\cite{b20}. The architecture consists of a generator and a discriminator (the ``Critic").  Conditioning information is introduced through a learnable embedding layer that maps discrete class labels into continuous vectors of fixed dimensionality. So given a class label, an embedding vector is obtained and combined with the input noise vector via element-wise multiplication. The generator is instantiated as a multilayer perceptron (MLP), which directly maps the conditioned latent vector to the image space.

\paragraph{FFT-based pattern guidance}\label{AA}
DF-ACBlurGAN employs a dynamic FFT-based analysis pipeline during generation for unit cell detection. Unlike conventional settings where frequency analysis is applied to clean, fully formed patterns, our estimation is performed directly on intermediate generator outputs. During early adversarial training, outputs often show incomplete repetition, blurred boundaries, and high-frequency artefacts, leading to unstable or spurious frequency responses (see Fig.~\ref{fig:qualitative_aeruginosa}).  To address this, our generated image is first normalized and transformed into the frequency domain using a two-dimensional discrete Fourier transform. The resulting magnitude spectrum is centred and projected onto the horizontal and vertical frequency axes by aggregating frequency responses across rows and columns. These projections emphasize the dominant frequencies associated with periodic repetition, as illustrated in Fig.~\ref{fig:fft_projection}. To distinguish dominant frequencies from potentially noisy generator outputs, dynamic thresholding is employed during peak detection. Rather than using a fixed global threshold, the peak selection threshold is defined relative to the distribution of frequency magnitudes:
\[
\tau = \min(\mathrm{Freq}) + \alpha \bigl(\max(\mathrm{Freq}) - \min(\mathrm{Freq})\bigr),
\]
where $\mathrm{Freq}$ denotes the projected frequency response and $\alpha$ controls sensitivity to prominent peaks. 

\begin{figure*}[t]
    \centering
    \includegraphics[width=\linewidth]{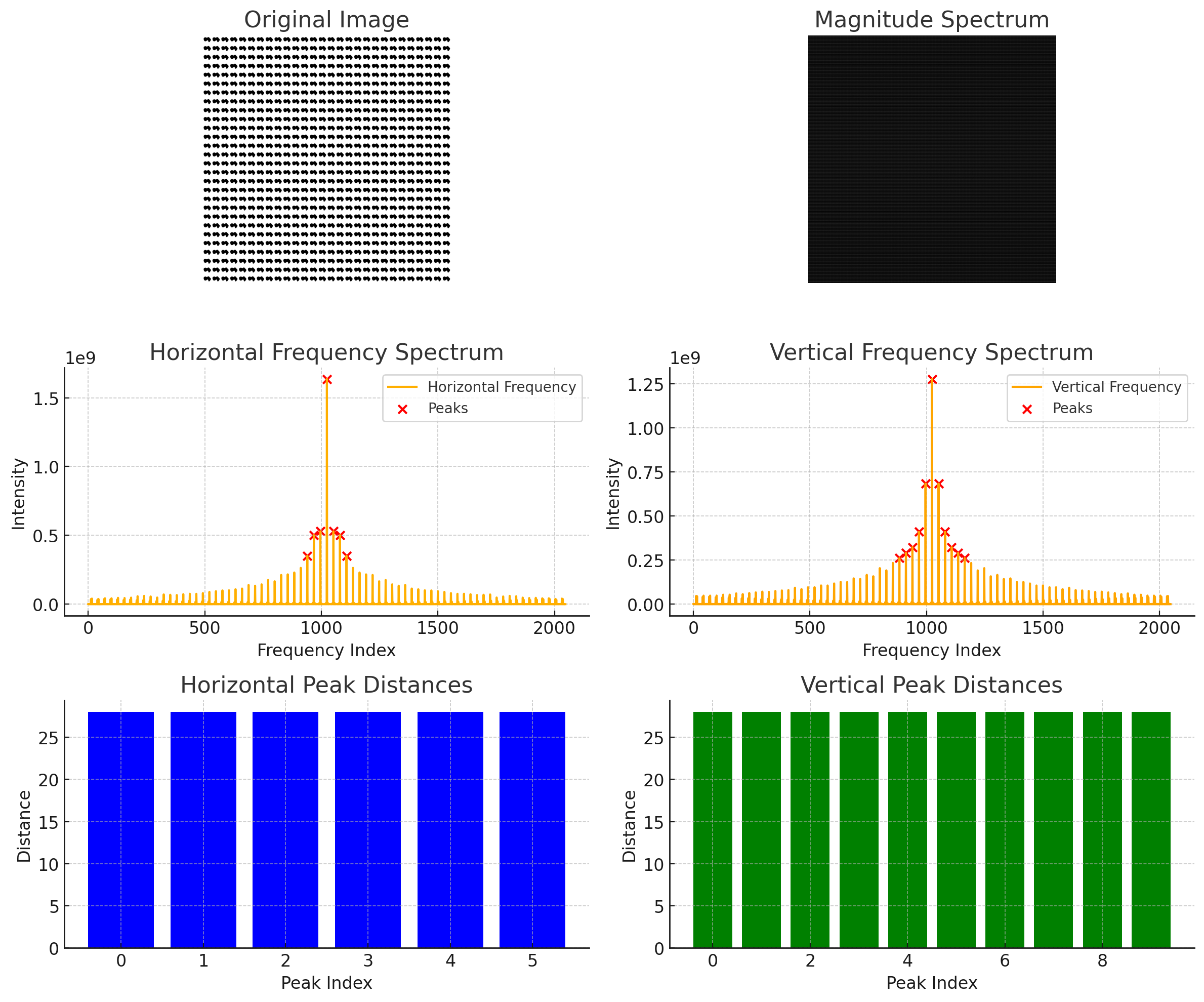}
    \caption{
    Illustration of FFT-based repetition analysis for patterned topographies.
    }
    \label{fig:fft_projection}
\end{figure*}

Additional consistency checks are applied during peak identification and spacing estimation to avoid false positive detections caused by weak or ambiguous frequency responses. Specifically, peaks are required to exceed the threshold and to appear consistently across neighbouring frequency bins; isolated peaks with irregular spacing are discarded. The distances between consecutive dominant peaks are then computed, and the most frequently occurring spacing is selected as the \emph{unit cell number}. To ensure compatibility with subsequent structure-aware operations, the estimated unit cell number is constrained to fit within the image size.

\paragraph{Adaptive Gaussian blurring}\label{AA}
DF-ACBlurGAN uses the unit cell number to drive an adaptive Gaussian blurring operation applied prior to loss computation. This addresses the tension between preserving sharp and well-defined local feature boundaries and suppressing high-frequency artefacts that can destabilise global repetition during adversarial training, or make the image very pixelated. 

Formally, the blur kernel size $k$ is computed as
\[
k = \left\lceil \frac{1}{10}
\min\!\left(\frac{H}{p_h},\,\frac{W}{p_w}\right)
\right\rceil,
\]
where $H$ and $W$ denote the image height and width, respectively, and
$p_h$ and $p_w$ represent the estimated pattern numbers along the vertical
and horizontal directions. 
In our experiments, images are square and exhibit isotropic repetition and therefore set $p_h = p_w = p$.
The kernel size is enforced to be an odd integer to maintain a symmetric Gaussian filter. The resulting blurred image is used in the subsequent reconstruction and structure-aware loss evaluation, providing a soft structure-aware constraint that stabilizes adversarial training without imposing explicit spatial constraints.

\paragraph{Unit-cell reconstruction}\label{AA}
To enforce consistency across repeated structural units, the generated image is partitioned into a regular grid of unit cells according to their estimated number. Each tile is treated as  a candidate realization of the underlying structural unit. A representative unit cell is obtained via pixel-wise aggregation across all tiles (e.g., majority voting for binary images), forming a consensus pattern that captures the dominant repeated structure. This reconstructed unit cell is then tiled back to the original image resolution, yielding a globally consistent reconstruction aligned with the inferred repetition layout.

\paragraph{Training objective and optimization}\label{AA}
DF-ACBlurGAN is trained using a weighted combination of adversarial, classification, and structure-aware regularization losses. These components jointly encourage visual realism, semantic consistency, and internal repetition coherence in the generated images.

The generator loss is defined as
\[
\mathcal{L}_G =
\lambda_W \mathcal{L}_W +
\lambda_{\mathrm{cls}} \mathcal{L}_{\mathrm{cls}} +
\lambda_{\mathrm{blur}} \mathcal{L}_{\mathrm{blur}} +
\lambda_{\mathrm{recon}} \mathcal{L}_{\mathrm{recon}},
\]
where $\mathcal{L}_W$ denotes the Wasserstein adversarial loss, $\mathcal{L}_{\mathrm{cls}}$ is the conditional classification loss, $\mathcal{L}_{\mathrm{blur}}$ penalizes the discrepancy between generated images and their adaptively blurred counterparts, and $\mathcal{L}_{\mathrm{recon}}$ enforces agreement with the unit-cell-based reconstruction.

To stabilize training, the reconstruction loss $\mathcal{L}_{\mathrm{recon}}$ is activated only after a predefined number of epochs, allowing the generator to first learn class-discriminative and local structural features.

The Critic is trained following the WGAN-GP formulation, with the loss
\[
\mathcal{L}_D =
\lambda_W \mathcal{L}_W^D +
\lambda_{\mathrm{cls}} \mathcal{L}_{\mathrm{cls}}^D,
\]
where a gradient penalty is applied to $\mathcal{L}_W^D$ to enforce the Lipschitz constraint. The full training procedure, including critic updates, FFT-based pattern estimation, adaptive blurring, and unit-cell reconstruction, is summarized in Algorithm~\ref{alg:df-acblurgan}.

\begin{figure}[t]
\centering
\includegraphics[width=\linewidth]{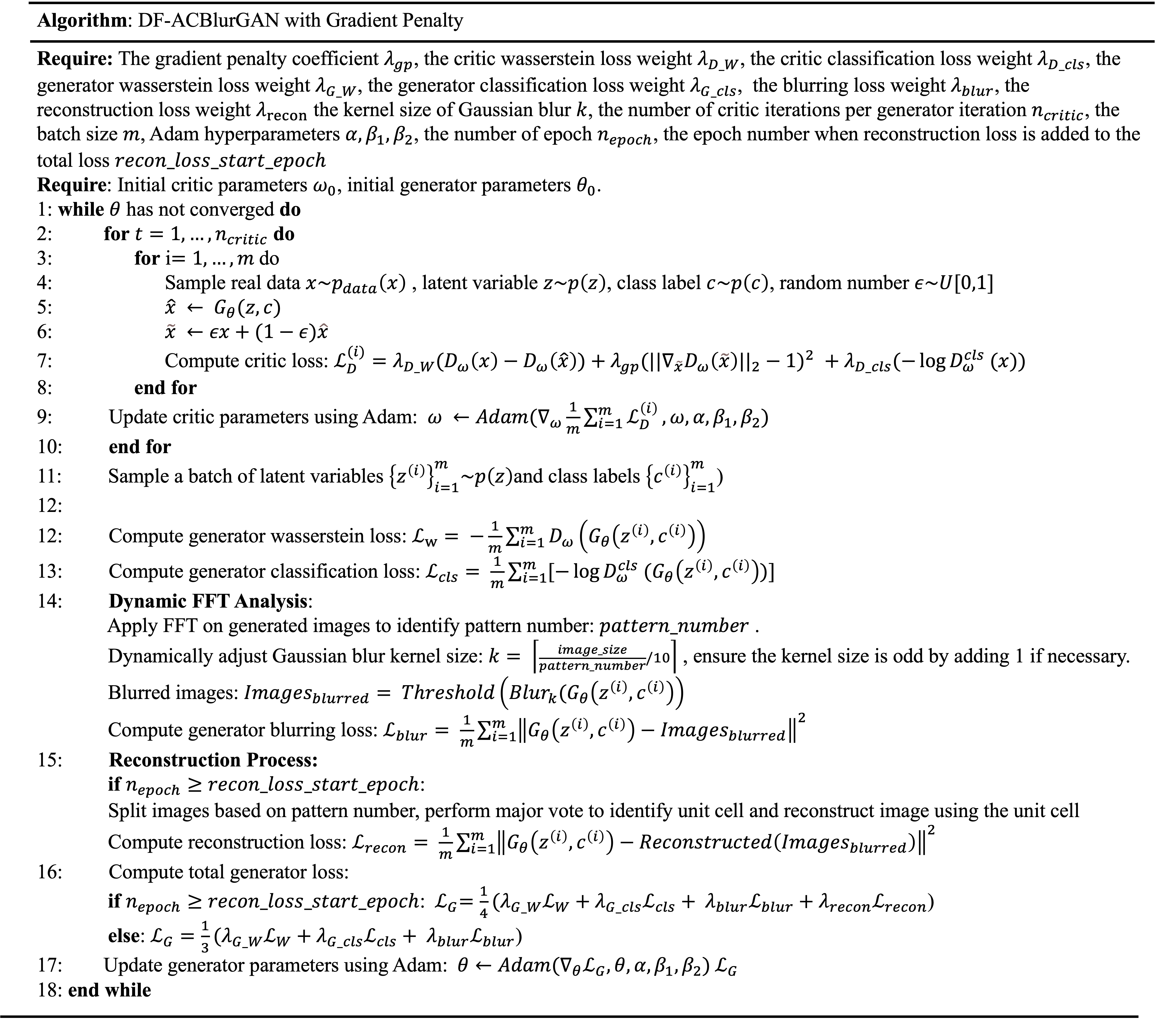}
\caption{Training procedure of DF-ACBlurGAN.}
\label{alg:df-acblurgan}
\end{figure}

\subsection{Evaluation Metrics and Validation Strategy}\label{AA}
Evaluating generative biomaterial topographies is challenging because no single metric can simultaneously capture visual plausibility, structural organisation, and biological relevance.
Visual inspection alone cannot assess internal repetition or structural consistency, while direct biological experiments are costly and impractical during model development.
We therefore adopt a complementary {\it in silico} evaluation strategy that combines
(i) standard perceptual metrics to assess overall visual realism and
(ii) domain-specific surrogate-based metrics to evaluate alignment with biologically meaningful feature representations.
These metrics are reported for DF-ACBlurGAN together with baseline and ablated variants for assessment of the contribution of individual architectural components.

\paragraph{Standard perceptual metrics.}
We report Fr\'echet Inception Distance (FID) and Inception Score (IS) as reference metrics, following common practice in generative modeling. However, as they are not specifically designed to capture repetitive microtopographical structures, they are treated as complementary indicators rather than primary measures of structural fidelity.

\paragraph{Evaluation via surrogate model predictions}
To evaluate generative quality within a problem context, we employ a pretrained ResNet-50 surrogate model trained on real topography images. 
Features extracted from this surrogate are used to define \emph{TopoFID}, a domain-specific Fr\'echet distance that measures the discrepancy between real and generated samples in the surrogate feature space, as well as a classifier-based Inception Score (IS\textsubscript{ResNet}).
The surrogate further serves as a downstream validation tool, where changes in predictive performance after incorporating selected generated samples are used to assess functional relevance, as reported in the Results section~\ref{sec:results}.

\section{Experimental Setup and Case Study Overview}
We evaluate DF-ACBlurGAN on three biomaterial topography datasets, covering bacterial biofilm formation (Pseudomonas (P) \textit{ aeruginosa} and Staphylococcus (S) \textit{aureus}) and a macrophage attachment--polarization task defined as $\log(\mathrm{M2/M1}) \times$ attachment~\cite{b4,b5,b7}. All topographies are represented as $2800\times2800$ binary images and paired with experimentally measured biological responses.
For all datasets, biological responses were discretized into three classes (low, moderate, high). After signal-to-noise filtering, each dataset was split into training and test subsets, with the test sets constructed to be class-balanced to enable fair evaluation. Dataset statistics are summarized in Table~\ref{tab:dataset_statistics}.
Quantitative evaluation using FID and IS is conducted on the \textit{P. aeruginosa} dataset, which provides the largest and most balanced distribution. In addition to DF-ACBlurGAN and its ablated variants, a vanilla conditional GAN without structure-aware components is included as a baseline to contextualise performance gains.
Surrogate-based evaluation and handcrafted structural descriptor analysis are then applied across all three datasets.

\begin{table}[t]
\centering
\caption{Dataset statistics after preprocessing and class discretization. All test sets are constructed to be class-balanced.}
\label{tab:dataset_statistics}
\begin{tabular}{lcccccc}
\toprule
\multirow{2}{*}{Dataset} & \multirow{2}{*}{Label} & \multicolumn{3}{c}{Number of Samples} \\
\cmidrule(lr){3-5}
 &  & Overall & Train & Test \\
\midrule
\multirow{3}{*}{\textit{P. aeruginosa}} 
 & 0 & 880 & 700 & 180 \\
 & 1 & 567 & 387 & 180 \\
 & 2 & 404 & 224 & 180 \\
\midrule
\multirow{3}{*}{\textit{S. aureus}} 
 & 0 & 619 & 469 & 150 \\
 & 1 & 1077 & 927 & 150 \\
 & 2 & 386 & 236 & 150 \\
\midrule
\multirow{3}{*}{Macrophage} 
 & 0 & 1488 & 1448 & 40 \\
 & 1 & 230 & 190 & 40 \\
 & 2 & 82 & 42 & 40 \\
\bottomrule
\end{tabular}
\end{table}

\section{Results}\label{sec:results}

\subsection{Quantitative and Qualitative Evaluation on P. \textit{aeruginosa}}

Table~\ref{tab:topofid_aeruginosa} reports FID, IS, and their domain-specific counterparts computed using a task-specific surrogate model (TopoFID and IS\textsubscript{ResNet}). While conventional FID and IS show limited separation between models, the surrogate-based metrics provide clearer discrimination. In particular, the full DF-ACBlurGAN achieves the lowest TopoFID and the highest IS\textsubscript{ResNet}, indicating that generated samples more closely match the structural statistics of real topographies in a biologically meaningful feature space. Ablation results further demonstrate the contribution of individual components. Removing FFT-based pattern guidance or unit-cell reconstruction consistently degrades TopoFID, highlighting the importance of explicit structural scale estimation and global alignment for repetitive pattern synthesis. Excluding adaptive blurring results in increased feature-space divergence, suggesting reduced suppression of high-frequency artifacts. In contrast, replacing the MLP generator with a convolutional backbone leads to a substantial deterioration in both TopoFID and IS\textsubscript{ResNet}, underscoring the difficulty of convolutional architectures in capturing long-range periodic structure.

\begin{table}[t]
\centering
\caption{Quantitative evaluation on \textit{P. aeruginosa}. Lower is better for FID/TopoFID, higher is better for IS.}
\label{tab:topofid_aeruginosa}
\footnotesize
\setlength{\tabcolsep}{3.5pt} 
\begin{tabularx}{\linewidth}{@{}Xcccc@{}}
\toprule
\textbf{Model} &
\textbf{FID}$\downarrow$ &
\textbf{IS}$\uparrow$ &
\textbf{TopoFID}$\downarrow$ &
\textbf{IS$_{\mathrm{Res}}$}$\uparrow$ \\
\midrule
ACVanillaGAN\_2800
& 246.673 & 1.821$\pm$0.113 & 349.989 & 1.571$\pm$0.091 \\
\midrule
w/o MLP Backbone (CNN)
& 473.936 & 1.760$\pm$0.048 & 1674.909 & 1.000$\pm$0.000 \\
w/o FFT Guidance
& 254.549 & 3.234$\pm$0.233 & 135.888 & 1.901$\pm$0.107 \\
w/o Adaptive Blur
& 252.803 & 3.271$\pm$0.427 & 104.928 & 1.984  $\pm$0.108 \\
w/o Unit-cell Reconstruction
& 264.518 & 3.044$\pm$0.224 & 96.278 & 1.927$\pm$0.173 \\
\midrule
\textbf{DF-ACBlurGAN (Full)}
& \textbf{246.859} & 3.082$\pm$0.252 & \textbf{86.690} & \textbf{2.107$\pm$0.163} \\
\bottomrule
\end{tabularx}
\end{table}

Fig.~\ref{fig:qualitative_aeruginosa} exemplifies how DF-ACBlurGAN produces spatially consistent and regularly repeated internal structures across both global views and zoomed-in regions in comparison to other other architectures used in our ablasion study. Without FFT guidance, repetition scales become unstable across the image, resulting in spatial drift and inconsistent pattern spacing. Removing adaptive blurring introduces localized high-frequency artifacts, while omitting unit-cell reconstruction disrupts global alignment despite locally plausible patterns. The convolutional baseline further fails to preserve coherent repetition, instead generating texture-like structures without clear periodic organization.

\begin{figure}[htbp]
\centering
\includegraphics[width=\linewidth]{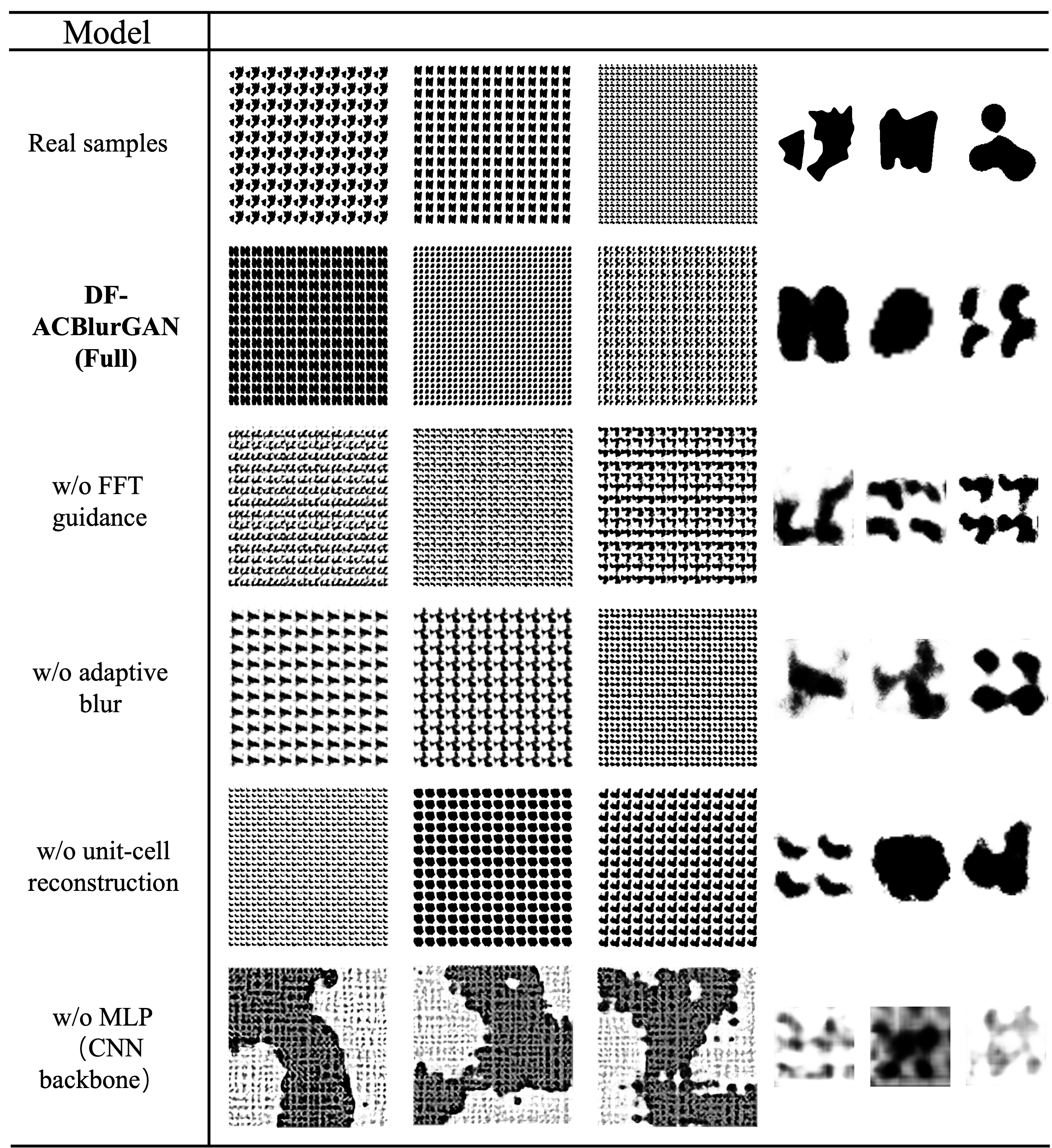}
\caption{Qualitative comparison of generated \textit{P. aeruginosa} topographies
under different model variants.}
\label{fig:qualitative_aeruginosa}
\end{figure}

Results demonstrate that the proposed structure-aware components act synergistically to enforce both local regularity and global consistency, enabling DF-ACBlurGAN to synthesize repetitive biomaterial topographies that better reflect the structural characteristics of experimentally fabricated surfaces.

Together, these results demonstrate that DF-ACBlurGAN preserves task-relevant structural statistics across diverse biological settings, providing structure-consistent generation beyond visual similarity alone.

\subsection{Functional Validation via Surrogate Model Performance}

We evaluate the functional usefulness of DF-ACBlurGAN-generated samples by measuring their impact on downstream surrogate model performance across three biological tasks.
For each dataset, a ResNet-50 surrogate was first trained on the original, imbalanced training set to establish a baseline.
A single round of synthetic augmentation was then performed using DF-ACBlurGAN, where generated samples were filtered using a single-Gaussian class-specific confidence model estimated via expectation--maximization (EM), with a fixed confidence threshold ($\alpha = 90\%$) and added to underrepresented classes to form a balanced training set.

As summarized in Table~\ref{tab:surrogate_all}, synthetic augmentation consistently improves surrogate performance across all datasets.
For the \textit{P.~aeruginosa} dataset, overall accuracy increased from 75.0\% to 80.2\%, with macro-averaged F1-score improving from 0.75 to 0.80.
Similar gains were observed for \textit{S.~aureus}, where accuracy improved from 73.8\% to 78.0\% and macro-F1 from 0.74 to 0.79.
Notably, for the macrophage log(M2/M1)$\times$attachment task, which exhibits severe class imbalance and high biological variability, augmentation led to a substantial increase in accuracy from 56.7\% to 71.7\% and macro-F1 from 0.53 to 0.71.

Across all tasks, improvements are most pronounced for previously underrepresented or intermediate-response classes, indicating that DF-ACBlurGAN-generated samples effectively enrich sparse regions of the feature space.
These results demonstrate that even a single iteration of carefully filtered synthetic augmentation can significantly enhance surrogate model generalization under diverse imbalance regimes.

\begin{table}[t]
\centering
\caption{Surrogate model performance before and after DF-ACBlurGAN-based synthetic augmentation across three biological datasets.}
\label{tab:surrogate_all}
\resizebox{\columnwidth}{!}{%
\begin{tabular}{lcccc}
\toprule
Dataset & Augmentation & Accuracy (\%) & Macro-F1 & Test Samples \\
\midrule
\textit{P.~aeruginosa} 
& Baseline   & 75.0 & 0.75 & 540 \\
& + DF-ACBlurGAN & \textbf{80.2} & \textbf{0.80} & 540 \\
\midrule
\textit{S.~aureus} 
& Baseline   & 73.8 & 0.74 & 450 \\
& + DF-ACBlurGAN & \textbf{78.0} & \textbf{0.79} & 450 \\
\midrule
Macrophage (log(M2/M1)$\times$Att.) 
& Baseline   & 56.7 & 0.53 & 120 \\
& + DF-ACBlurGAN & \textbf{71.7} & \textbf{0.71} & 120 \\
\bottomrule
\end{tabular}
}
\end{table}

\subsection{Discussion}

DF-ACBlurGAN for synthesising images with internally repeated and periodic structures under class imbalance has produced satisfactory outcomes. A central contribution lies in the use of FFT-based for unit cell detection during training. Rather than relying on fixed architectural priors or post-hoc frequency regularisation, DF-ACBlurGAN estimates repetition scale directly from intermediate generator outputs and feeds this information back into downstream refinement operations. The choice of an MLP-based generator further distinguishes DF-ACBlurGAN from convolution-based generative architectures. While CNNs are effective at capturing local texture statistics, they fail to capture long-range periodic dependencies in large images. By directly mapping conditioned latent vectors to image space, the MLP backbone facilitates global structural coordination, which is subsequently stabilised through structure-aware losses. Ablation results confirm that this design choice is the best suited for maintaining coherent repetition. Adaptive Gaussian blurring and unit-cell reconstruction, commonly applied as pre- or post-processing steps, are integrated directly into the learning objective as scale-aware regularisers. Experiments show that blurring suppresses high-frequency artefacts in a repetition-aligned manner, while unit-cell reconstruction enforces consistency across repeated structural units. These components act synergistically, constraining both local smoothness and global organisation without imposing rigid geometric templates, contributing to stable training and improved structural fidelity.

\section{Conclusions}

DF-ACBlurGAN is a structure-aware conditional generative architecture for synthesising design images with periodic repetitive patterns under class imbalance. By integrating FFT-based repetition estimation, adaptive blurring, and unit-cell reconstruction into adversarial training, the approach enforces structural consistency during generation. Experiments across three biological tasks demonstrate structure-consistent desigs, validated through quantitative metrics and surrogate-based evaluation. In addition, EM-based filtering enables effective one-shot data augmentation, improving downstream surrogate performance. Beyond biomaterial surface design, the framework generalises to other generative design problems involving repeated patterns, including photonic materials, metamaterials, microfabrication layouts, and architectural surface patterns. The underlying principles of frequency-guided structural inference and reconstruction-based regularisation are model-agnostic and transferable to alternative generative paradigms such as diffusion models and normalising flows. Several limitations remain. Discretised class labels derived from continuous biological responses introduce semantic ambiguity, particularly for intermediate regimes, motivating future exploration of regression-based or continuous conditioning. While FFT-based estimation captures global periodicity, more complex hierarchical repetition may require multi-scale or wavelet-based extensions. Finally, integrating iterative experimental feedback or physics-informed and fabrication constraints would further strengthen closed-loop design capabilities.

\vspace{12pt}

\end{document}